\documentclass[journal]{IEEEtran}

\usepackage{amsmath, amssymb, amsfonts}
\usepackage{graphicx}
\usepackage{graphicx}
\usepackage{booktabs}
\usepackage[bookmarks=false,
            colorlinks=true,
            citecolor=black,   
            linkcolor=black, 
            urlcolor=black]
            {hyperref}

\usepackage{cite}
\usepackage{algorithm}
\usepackage[caption=false,font=scriptsize]{subfig}
\usepackage{algorithmic}
\usepackage{url}
\usepackage{booktabs}

\begin{document}
\title{Federated Multi-Agent Reinforcement Learning for Privacy-Preserving and Energy-Aware Resource Management in 6G Edge Networks}

\author{%
\IEEEauthorblockN{Francisco J. E. N. Andong\IEEEauthorrefmark{1}, Qi Min\IEEEauthorrefmark{2}}
\IEEEauthorblockA{\IEEEauthorrefmark{1}\IEEEauthorrefmark{2}Department of Electronics and information Engineering, Northwestern Polytechnical University (NWPU), Xi'an, China\\
Emails: esonofranciscojavier@mail.nwpu.edu.cn, drqimin@nwpu.edu.cn}
}

\maketitle

\begin{abstract}
As sixth-generation (6G) networks move toward ultra-dense, intelligent edge environments, efficient resource management under stringent privacy, mobility, and energy constraints becomes critical. This paper introduces a novel Federated Multi-Agent Reinforcement Learning (Fed-MARL) framework that incorporates cross-layer orchestration of both the MAC layer and application layer for energy-efficient, privacy-preserving, and real-time resource management across heterogeneous edge devices. Each agent uses a Deep Recurrent Q-Network (DRQN) to learn decentralized policies for task offloading, spectrum access, and CPU energy adaptation based on local observations (e.g., queue length, energy, CPU usage, and mobility). To protect privacy, we introduce a secure aggregation protocol based on elliptic-curve Diffie–Hellman key exchange, which ensures accurate model updates without exposing raw data to semi-honest adversaries. We formulate the resource management problem as a partially observable multi-agent Markov decision process (POMMDP) with a multi-objective reward function that jointly optimizes latency, energy efficiency, spectral efficiency, fairness, and reliability under 6G-specific service requirements such as URLLC, eMBB, and mMTC. Simulation results demonstrate that Fed-MARL outperforms centralized MARL and heuristic baselines in task success rate, latency, energy efficiency, and fairness, while ensuring robust privacy protection and scalability in dynamic, resource-constrained 6G edge networks.
\end{abstract}

\section{Introduction}
Sixth-generation (6G) wireless networks are poised to transform communication systems by enabling ultra-dense connectivity, low-latency services, and intelligent edge processing capabilities~\cite{ref1}. These advances are critical for emerging applications such as autonomous driving, augmented reality, and massive Internet of Things (IoT) deployments, each imposing diverse and stringent quality-of-service (QoS) requirements~\cite{ref2,ref3}. Efficiently meeting these demands requires decentralized, real-time resource management frameworks capable of operating in highly dynamic, interference-prone, and energy-constrained environments under strict privacy conditions.

Traditional centralized resource management architectures, which depend on global network knowledge for task offloading, spectrum allocation, and computational scheduling, face significant limitations in 6G contexts~\cite{ref4,ref5}. These include scalability bottlenecks, latency, communication overhead, and privacy concerns, particularly when raw user data must be aggregated~\cite{ref6}. Moreover, node mobility and wireless channel variability further degrade the feasibility of centralized control in ultra-dense edge networks~\cite{ref6}. As a result, decentralized intelligence that enables autonomous agents to make context-aware decisions based on local observations has emerged as a critical design requirement~\cite{ref7}.

Multi-agent reinforcement learning (MARL) offers a promising paradigm for distributed decision-making, allowing agents to learn cooperative policies through environmental interaction and limited coordination~\cite{ref8,ref9}. However, many existing MARL solutions assume centralized training or full observability, which are incompatible with the partial observability, dynamic topologies, and privacy constraints of 6G edge networks~\cite{ref10,ref11,ref12,ref13}. Additionally, most frameworks focus on optimizing a single metric (e.g., latency or throughput), overlooking essential trade-offs involving fairness, energy efficiency, and spectral utilization~\cite{ref14,ref15}.

Federated learning (FL) has recently emerged as a privacy-preserving alternative that supports collaborative training through local computation and periodic parameter aggregation, without exposing raw data~\cite{ref16,ref17}. Combining FL with MARL yields the federated multi-agent reinforcement learning (Fed-MARL) approach, which facilitates distributed policy learning from partial observations while enabling communication-efficient synchronization~\cite{ref18,ref19}. Nonetheless, standard FL is susceptible to gradient inversion and related inference attacks during model update exchanges~\cite{ref20,ref21}.

To enhance privacy, we propose a Fed-MARL framework that incorporates a secure aggregation protocol using cryptographic pairwise masking and elliptic-curve Diffie–Hellman (ECDH) key exchange. Assuming synchronous updates, this mechanism guarantees that only the aggregated model is revealed, preserving the confidentiality of individual gradients and strengthening privacy in dynamic 6G edge environments~\cite{ref22}.

Despite its potential, current MARL approaches remain limited by their assumptions of global observability and centralized coordination~\cite{ref23}. Moreover, many solutions neglect multi-objective optimization, often focusing narrowly on either MAC-layer or application-layer policies~\cite{ref24,ref25,ref26}, and rarely consider realistic scenarios involving mobility, traffic heterogeneity, and energy constraints~\cite{ref27}. Addressing these limitations requires a scalable, adaptive, and privacy-preserving learning framework that supports decentralized control under partial observability.

In this paper, we propose a novel Fed-MARL framework for cross-layer, multi-objective resource management in 6G edge networks. Our main contributions are summarized as follows:
\begin{itemize}
   \item We propose a novel Fed-MARL framework that enables decentralized, cross-layer task offloading, spectrum access, and CPU energy adaptation for 6G edge networks, leveraging privacy-preserving federated learning and multi-agent reinforcement learning.
    
    \item We introduce a cross-layer coordination mechanism that jointly manages application-layer decisions (e.g., task offloading) and MAC-layer decisions (e.g., spectrum access and energy consumption), addressing dynamic resource constraints and heterogeneous service types in 6G.

      \item We introduce a novel adaptation of the secure aggregation protocol from Google’s framework to federated multi-agent reinforcement learning, enabling model privacy in dynamic edge environments. This adaptation combines elliptic-curve Diffie–Hellman key exchange and symmetric masking, ensuring strong privacy protection under semi-honest adversaries.
    
    \item We model the resource management problem as a partially observable multi-agent Markov decision process (POMMDP), incorporating multi-objective optimization to jointly optimize latency, energy efficiency, spectral efficiency, fairness, and reliability under real-world mobility and traffic conditions in 6G networks.
    
    \item We demonstrate, through extensive simulations, that Fed-MARL significantly outperforms centralized MARL and heuristic baselines in terms of task success rate, latency, energy efficiency, and fairness, while maintaining strong privacy protections and reducing communication overhead in dynamic, resource-constrained 6G edge environments.
\end{itemize}

The paper is structured as follows: Section~\hyperref[sec:related_work]{II} reviews related work in multi-agent reinforcement learning, federated learning, and wireless network resource management. Section~\hyperref[sec:system_architecture_and_problem_formulation]{III} introduces the system model and formulates the resource coordination problem. Section~\hyperref[sec:proposed_method]{IV} outlines the proposed Fed-MARL framework, including its training process and reward design. Section~\hyperref[sec:experimental_evaluation]{V} presents the simulation setup and performance evaluation, Section~\hyperref[sec:discussion]{VI} discusses trade-offs between performance and complexity in 6G. Finally, Section~\hyperref[sec:conclusion]{VII} concludes the paper and suggests future research directions.

\section{Related Work}
\label{sec:related_work}

\subsection{Resource Management in 6G and Edge Networks}
Resource management is a foundational challenge in 6G wireless systems, driven by the need for ultra-reliable low-latency communication in ultra-dense, dynamic environments. Gupta et al.~\cite{ref28} highlighted the complexities of spectrum reuse and interference management in device-to-device (D2D) settings, while Kamath~\cite{ref29} discussed orchestration inefficiencies in centralized 5G control architectures. These approaches often assume global network visibility, which is impractical in decentralized and privacy-sensitive edge environments. Consequently, recent research has shifted toward decentralized and context-aware schemes that better accommodate the mobility, heterogeneity, and privacy constraints expected in 6G.

\subsection{Multi-Agent Reinforcement Learning in Wireless and Edge Environments}
To support decentralized intelligence, multi-agent reinforcement learning (MARL) has gained traction as a method for adaptive decision-making. Liu et al.~\cite{ref30} applied MARL for spectrum access and task offloading in IoT-edge systems but relied on simplified settings and limited scalability. Zhang et al.~\cite{ref31} addressed scalability using consensus-based decentralized MARL; however, their methods struggle with partial observability and dynamic network conditions, and they lack privacy and energy-awareness. In contrast, the proposed Fed-MARL framework addresses these limitations by employing Deep Recurrent Q-Networks (DRQN) for partial observability, secure federated aggregation for privacy, and energy-aware decision-making suitable for 6G edge scenarios.

\subsection{Federated Learning for Distributed Decision-Making}
Federated learning (FL) enables collaborative model training across devices without exchanging raw data, thus enhancing privacy. Sharma et al.~\cite{ref32} proposed a general FL architecture for edge systems, and Elbir et al.~\cite{ref33} extended FL to vehicular networks. However, most FL-based studies neglect integration with MARL, overlook multi-objective optimization, and often assume static conditions, limiting their applicability to mobile, dynamic 6G environments where learning must be robust to partial information and rapid state changes.

\subsection{Energy-Aware and Cross-Layer Optimization}
Energy efficiency is critical in edge networks where devices operate under strict energy budgets. Sada et al.~\cite{ref34} introduced a reinforcement learning-based offloading method balancing energy and latency, while Lu et al.~\cite{ref35} surveyed RL techniques for cross-layer security at the PHY and MAC layers. Yet, many of these efforts focus on isolated layers or objectives and often rely on centralized control. They rarely integrate energy, latency, and reliability into a unified framework capable of operating under full decentralization.

\subsection{Federated Reinforcement Learning for Edge Intelligence}
Recent advances have attempted to unify FL and RL to enable decentralized learning with privacy guarantees. The FAuNO framework~\cite{ref36} uses a semi-asynchronous federated actor–critic method to support partial observability in task offloading but focuses on single-layer optimization and lacks robust cryptographic privacy mechanisms. Other efforts, such as those by Zarandi and Tabassum~\cite{ref37} and Noman et al.~\cite{ref38}, show promise but often omit fairness, joint optimization, or strong protection against inference attacks. The proposed Fed-MARL framework fills these gaps by integrating DRQN-based temporal modeling, cryptographic aggregation, and multi-layer optimization across latency, energy, spectral efficiency, and fairness.

\subsection{Research Gaps and Motivation}
Despite advancements across MARL, FL, and energy-aware computing, most prior works treat these areas in isolation or under idealized assumptions such as centralized coordination, full observability, or static environments. These assumptions are incompatible with real-world 6G edge deployments characterized by mobility, energy constraints, and privacy risks. This paper addresses these limitations by introducing a unified Fed-MARL framework that combines multi-agent learning, secure federated training, partial observability, and multi-objective cross-layer optimization, offering a practical solution for intelligent resource management in next-generation 6G networks.

\section{System Architecture and Problem Formulation}
\label{sec:system_architecture_and_problem_formulation}

\subsection{System Model}
\label{sec:section_III_A}
\begin{figure}[h]
  \centering
  \includegraphics[width=0.45\textwidth]{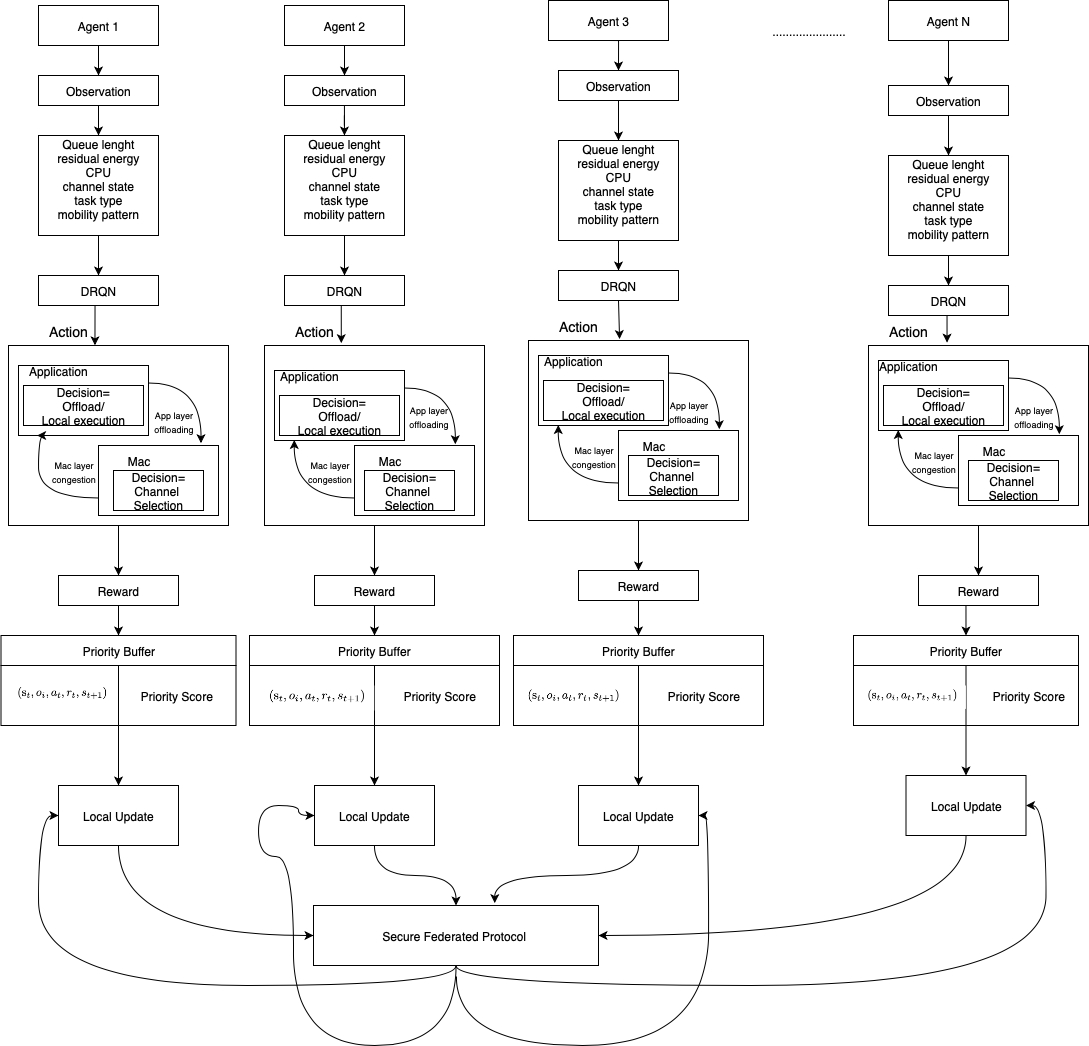}
  \caption{System Architecture of FERMI-6G.}
  \label{fig:image1}
\end{figure}

We consider a federated multi-agent wireless edge computing environment reflecting ultra-dense, decentralized, and dynamic 6G conditions. As shown in Fig. \ref{fig:image1}, the system consists of a set of agents $\mathcal{N} = \{1,2,\dots,N\}$, each representing an edge node capable of generating and processing computational tasks while contending for a shared wireless spectrum. These agents operate under partial observability, without centralized coordination, and must make real-time decisions across the application, MAC, and CPU layers to manage task execution, communication, and energy consumption.

At each decision interval $t$, agent $i$ selects an action vector:
\begin{equation}
\mathbf{a}_i(t) = \big(a_i^{\text{app}}(t), a_i^{\text{mac}}(t), a_i^{\text{cpu}}(t)\big),
\end{equation}
where $a_i^{\text{app}}(t) \in \{0,1\}$ determines whether a task is processed locally or offloaded, $a_i^{\text{mac}}(t) \in \{1,\dots,k\}$ specifies the transmission channel, and $a_i^{\text{cpu}}(t) \in \mathbb{R}$ denotes the CPU frequency for local execution. These actions are tightly coupled and must be jointly optimized for latency, energy efficiency, and fairness. The system supports multiple traffic types such as  Ultra-Reliable Low Latency Communications (URLLC), Enhanced Mobile Broadband (eMBB), and Massive Machine Type Communications (mMTC) with distinct QoS requirements influencing agent behavior.

\subsection{POMMDP-Based Environment Formulation}
\label{sec:section_III_B}
The learning environment is modeled as a partially observable multi-agent Markov decision process (POMMDP):
\begin{equation}
\mathcal{M} = \langle \mathcal{S}, \mathcal{A}, \mathcal{T}, \mathcal{O}, \mathcal{R}, \gamma \rangle,
\end{equation}
where $\mathcal{S}$ represents the global state including queue lengths, energy levels, channel states, task types, and mobility. Agent $i$ observes a local state:
\begin{equation}
o_i(t) = \{q_i(t), E_i(t), h_i(t), \lambda_i(t), c_i(t)\},
\end{equation}
with queue status $q_i(t)$, residual energy $E_i(t)$, channel gain $h_i(t)$, task type $\lambda_i(t)$, and CPU usage $c_i(t)$. The environment evolves stochastically according to the transition function $\mathcal{T}$, influenced by agent actions and external dynamics such as interference and mobility. Agents learn policies $\pi_i(o_i)$ through reinforcement under uncertainty.

\subsection{Agent Decision and Interaction}
\label{sec:section_III_C}
Each agent independently learns a policy using a Deep Recurrent Q-Network (DRQN) with LSTM to handle partial observability and temporal dependencies. Observations are mapped to action vectors $\mathbf{a}_i(t)$ for cross-layer decision-making across offloading, channel access, and CPU scaling. 

Agents store temporal sequences in a prioritized experience replay buffer for backpropagation through time (BPTT), enhancing stability in non-stationary environments. Training occurs locally, but agents synchronize periodically via secure federated aggregation using elliptic-curve Diffie–Hellman (ECDH) and symmetric AES masking. Pairwise noise masks ensure that the global model is recovered without exposing individual contributions.

\subsection{Multi-Objective Reward Function}
\label{sec:section_III_D}

The reward function relies on several key variables that capture task performance and system efficiency. Let $L$ denote the task latency, including processing, queuing, and transmission times. $D$ is the task deadline, while $R_i$ measures the reliability of task $i$ as a rolling average of successful completions. Fairness across MAC channels is captured by $F$ using hybrid's index, $SE$ represents the spectral efficiency defined as throughput per unit bandwidth, and $EE$ denotes energy efficiency measured as throughput per unit energy. The MAC success rate is indicated by $MAC_{\text{succ}}$. Each metric is associated with a task-specific weight, namely $w_L$, $w_E$, $w_F$, $w_{SE}$, $w_R$, $w_{EE}$, and $w_{MAC}$, corresponding to latency, energy, fairness, spectral efficiency, reliability, energy efficiency, and MAC success, respectively.

To ensure comparability across tasks, the key metrics are normalized. Latency tracks the time between task initiation and completion, which is crucial for time-sensitive applications like URLLC. The normalized latency for task $i$ at time $t$ is defined as
\begin{equation}
\tilde{L}_i(t) = \frac{L_i(t)}{D_i},
\end{equation}
where $\tilde{L}_i(t)$ is the normalized latency, $L_i(t)$ is the latency for task $i$ at time $t$, and $D_i$ is the task deadline. Energy consumed per task, including transmission and computation, is defined as

\begin{equation}
 E_i(t)= \frac{\sum_{t=1}^{T} \left(E_\text{tx}^{(i)}(t) + E_\text{comp}^{(i)}(t)\right)}{N_\text{tasks}}
\quad \text{[J/task]},
\end{equation}
where $E_i(t)$ is agent \(i\)'s energy per task and \(N_\text{tasks}\) is the total number of tasks. This metric encourages agents to reduce energy usage, critical for battery-constrained edge devices. The normalized energy consumption is
\begin{equation}
\tilde{E}_i(t) = \frac{E_i(t)}{E_{\text{max}}},
\end{equation}
where $\tilde{E}_i(t)$ is the normalized energy, $E_i(t)$ is the energy consumed by task $i$ at time $t$, and $E_{\text{max}}$ is the maximum energy capacity. Fairness across agents is measured using a combination of Jain's index and channel access entropy. 
Suppose there are $N$ agents, and let $x_i(t)$ denote the resource allocation (e.g., throughput, successful MAC attempts, or normalized reward) of agent $i$ at time $t$. 
Jain’s fairness at time $t$ is given by
\begin{equation}
F_{\text{Jain}}(t) = \frac{\left( \sum_{i=1}^{N} x_i(t) \right)^2}{N \sum_{i=1}^{N} x_i(t)^2}, \quad F_{\text{Jain}}(t) \in \left[\frac{1}{N}, 1\right].
\end{equation}

Suppose each agent $i$ can choose among $M$ channels. Let $p_i^c$ denote the probability that agent $i$ selects channel $c$ over a recent window of $T_w$ time steps. The normalized channel access entropy of agent $i$ is
\begin{equation}
\tilde{H}_i = - \frac{1}{\log_2 M} \sum_{c=1}^{M} p_i^c \log_2 p_i^c, \quad \tilde{H}_i \in [0,1],
\end{equation}
and the average channel access entropy across all agents is
\begin{equation}
H_{\text{avg}} = \frac{1}{N} \sum_{i=1}^{N} \tilde{H}_i.
\end{equation}
Finally, the hybrid fairness index used in the reward is
\begin{equation}
F_i(t) = \beta_{\text{jain}} F_{\text{Jain}}(t) + \beta_{\text{entropy}} H_{\text{avg}},
\end{equation}
where $\beta_{\text{jain}}$ and $\beta_{\text{entropy}}$ are tunable weights that balance instantaneous fairness and temporal channel diversity. Reliability is measured by the moving average of task success rates over time, providing an indication of the network’s long-term stability in completing tasks without failure. The Task Success Rate (TSR) quantifies the percentage of tasks successfully offloaded and completed within deadlines. The reliability of task $i$ is calculated as a rolling average of successes, 

\begin{equation}
R_i(t) = \frac{1}{n} \sum_{j=1}^{n} \text{success}_j,
\end{equation}
where $R_i(t)$ represents the reliability. The Spectral Efficiency (SE) metric quantifies how effectively data is transmitted per unit of spectrum, which reflects the system’s ability to maximize bandwidth usage in dense environments. The normalized spectral efficiency is
\begin{equation}
\tilde{SE}_i(t) = \max\big(SE_i(t), 0.01\big),
\end{equation}
where $\tilde{SE}_i(t)$ is the normalized spectral efficiency. Energy efficiency, the ratio of throughput to energy, is

\begin{equation}
EE_i(t) = \frac{\text{Throughput}_i(t)}{E_\text{total}(t)}
\quad \text{[bits/J]},
\end{equation}
capturing the trade-off between performance and energy, motivating agents to maximize throughput while minimizing consumption. Finally,  the MAC Success Rate indicates the percentage of successful MAC transmissions, revealing how well the system handles communication despite potential congestion. MAC success rate is
\begin{equation}
MAC_{\text{rate},i}(t) = \frac{\text{successful MAC attempts}}{\text{total MAC attempts}},
\end{equation}
where $MAC_{\text{rate},i}(t)$ denotes the MAC success rate for task $i$ at time $t$.

Dynamic penalties are applied to stabilize learning based on latency and energy:

\begin{align}
P_{\text{dyn}} &=
\begin{cases}
2, & L_i(t) > 2~\text{sec},\\
1, & \text{otherwise},
\end{cases} \\
P_{\text{energy}} &=
\begin{cases}
2, & \text{remaining energy} < \text{threshold},\\
1, & \text{otherwise}.
\end{cases}
\end{align}
The total reward for agent $i$ at time $t$ is then computed as:

\begin{align}
r_i(t) &= w_L \, \tilde{L}_i^{-1}(t) \, P_{\text{dyn}}
        + w_E \, \tilde{E}_i^{-1}(t) \, P_{\text{energy}} \notag\\
       &\quad + w_F \, F_i(t)
        + w_R \, R_i(t) \notag\\
       &\quad + w_{SE} \, \tilde{SE}_i(t)
        + w_{EE} \, EE_i(t)
        + w_{MAC} \, MAC_{\text{rate},i}(t),
\end{align}
where $\tilde{L}_i^{-1}(t)$ and $\tilde{E}_i^{-1}(t)$ are the inverse normalized latency and energy, respectively, ensuring that lower latency and energy consumption correspond to higher rewards. $P_{\text{dyn}}$ and $P_{\text{energy}}$ are the dynamic penalties applied based on latency and remaining energy.

To facilitate modular analysis and cross-layer learning, we decompose the total reward into application-layer and MAC-layer components. The application-layer reward, which focuses on computation and latency trade-offs, is given by

\begin{equation}
r_i^{\text{app}}(t) = -\alpha \cdot \tilde{L}_i^{-1}(t) \cdot P_{\text{dyn}} - \beta \cdot \tilde{E}_{\text{comp},i}^{-1}(t) \cdot P_{\text{energy}},
\end{equation}
where $\tilde{L}_i^{-1}(t)$ is the inverse normalized latency and $\tilde{E}_{\text{comp},i}^{-1}(t)$ is the inverse normalized energy consumption related to local computation. The MAC-layer reward, which focuses on transmission energy and fairness, is expressed as

\begin{equation}
r_i^{\text{MAC}}(t) = -\beta \cdot \tilde{E}_{\text{trans},i}^{-1}(t) \cdot P_{\text{energy}} + \gamma \cdot F_i(t),
\end{equation}
where $\tilde{E}_{\text{trans},i}^{-1}(t)$ is the inverse normalized energy consumption related to transmission and $F_i(t)$ represents fairness, captured using the hybrid index. Together, these components allow the reward function to consider both computation and communication aspects of task execution while incorporating dynamic penalties for latency and energy.

The total reward for agent $i$ at time $t$ is the sum of the application layer reward, the MAC layer reward, and additional metrics:

\begin{equation}
r_i(t) = r_i^{\text{app}}(t) + r_i^{\text{MAC}}(t) + \lambda \cdot \Omega_i(t),
\end{equation}
where $\Omega_i(t)$ represents additional factors such as reliability, spectral efficiency, or other task-specific metrics, and $\lambda$ is the weight applied to these additional metrics.

This structure allows for layer-specific emphasis during training while promoting end-to-end optimization. It also encourages agents to understand trade-offs between local execution and offloading, as well as how MAC-layer decisions impact both latency and fairness under dynamic network conditions.

\subsection{Federated Learning Protocol}
\label{sec:section_III_E}

Agents synchronize their Deep Recurrent Q-Network (DRQN) models every $K$ episodes using a secure federated aggregation protocol. After local training, each agent encrypts its model update using symmetric keys derived via Elliptic Curve Diffie-Hellman (ECDH) key exchange, combined with additive masking~\cite{ref21}. 

Let $w_i$ denote the local model parameters (or update) of agent $i$, and let $m_{ij}$ be the shared pseudorandom mask generated between agents $i$ and $j$, derived from their shared symmetric key. The total mask $M_i$ applied by agent $i$ is defined as:
\begin{equation}
M_i = \sum_{j>i} m_{ij} - \sum_{j<i} m_{ji} .
\end{equation}

The masked model update transmitted by agent $i$ is then:
\begin{equation}
\tilde{w}_i = w_i + M_i .
\end{equation}

The aggregator computes the average over all masked updates:
\begin{equation}
w = \frac{1}{N} \sum_{i=1}^N \tilde{w}_i = \frac{1}{N} \sum_{i=1}^N (w_i + M_i) .
\end{equation}

Observe that each pairwise mask $m_{ij}$ appears exactly once with a positive sign and once with a negative sign across the sum $\sum_{i=1}^N M_i$, i.e.,
\begin{equation}
\sum_{i=1}^N M_i = \sum_{i=1}^N \left( \sum_{j>i} m_{ij} - \sum_{j<i} m_{ji} \right) = 0 .
\end{equation}

Therefore, the masks cancel out in the aggregate and the final aggregation becomes:
\begin{equation}
w = \frac{1}{N} \sum_{i=1}^N w_i.
\end{equation}

This protocol ensures privacy of individual updates through masking while allowing the server to compute the exact global average. The aggregated model is broadcast to all agents, providing fairness, consistency, and scalability, with efficient communication and robustness in mobile networks.

\subsection{Optimization Problem}
\label{sec:section_III_F}

The global objective of Fed-MARL is to learn decentralized policies $\{\pi_i\}_{i=1}^N$ that maximize long-term cumulative rewards under constraints:
\begin{equation}
\begin{aligned}
\max_{\pi} \ & \mathbb{E}\Bigg[\sum_{t=0}^{\infty} \gamma^t r_i(t) \Bigg] \\
\text{s.t. } & E_i(t) \ge E_i^{\min}, \quad \forall i,t, \\
& o_i(t) \subseteq s(t), \quad \text{(Partial Observability)} \\
& \text{Communication cost} \le C_{\max}.
\end{aligned}
\end{equation}

This formulation captures the challenge of learning scalable and privacy-preserving control policies for edge agents in resource-constrained, dynamic 6G environments.

\section{Proposed Method}
\label{sec:proposed_method}

\subsection{Overview of the FERMI-6G Framework} 
We propose FERMI-6G (Federated Energy-aware Reinforcement for Multi-agent Intelligence in 6G), a novel federated multi-agent reinforcement learning (Fed-MARL) framework for decentralized, privacy-preserving, and energy-efficient resource management in ultra-dense 6G edge networks. Building on the system formulation in Section~\hyperref[sec:system_architecture_and_problem_formulation]{III}, FERMI-6G introduces four key innovations:

Temporal Deep Reinforcement Learning: Each agent employs a Deep Recurrent Q-Network (DRQN) with LSTM layers to capture temporal dependencies and manage uncertainty under partial observability (see Section~\hyperref[sec:section_III_C]{III-C}).

Cross-layer Composite Action Space: As defined in Section~\hyperref[sec:section_III_A]{III-A}), agents select a joint action vector 
\begin{equation}
a_i(t) = (a_i^{\text{app}}(t), a_i^{\text{MAC}}(t), a_i^{\text{CPU}}(t)),
\end{equation} 
enabling integrated optimization of task offloading, spectrum access, and CPU scaling.

Secure Federated Synchronization: Agents periodically synchronize their models via a privacy-preserving aggregation scheme based on elliptic-curve Diffie–Hellman (ECDH) and AES masking, protecting sensitive updates during coordination (see Section~\hyperref[sec:section_III_E]{III-E} ).

Adaptive Multi-Objective Optimization: FERMI-6G employs a dynamic reward function defined in Section~\hyperref[sec:section_III_D]{III-D}, which balances latency, energy consumption, fairness, spectral efficiency, and reliability. The weights of each reward component are adjusted in real time using moving averages of system-level metrics, enabling agents to adapt to diverse traffic types (URLLC, eMBB, mMTC), mobility patterns, and energy constraints.

These integrated mechanisms enable FERMI-6G to operate effectively in multi-agent 6G environments characterized by partial observability, mobility, energy constraints, and dynamic spectrum access.

\subsection{Algorithmic Structure of FERMI-6G}
FERMI-6G operates in episodic, synchronous rounds. During each episode, every agent independently interacts with its local environment, maintains its internal state via an LSTM, and updates its DRQN policy using prioritized experience replay. The reward function $r_i(t)$ is computed as defined in Section~\hyperref[sec:section_III_D]{III-D}, combining latency, energy, and fairness components with adaptive weights.

Below is the detailed pseudocode representation of FERMI-6G's federated multi-agent learning workflow:

\begin{algorithm}[H]
\caption{FERMI-6G Sub-Algorithm 1A: Local Agent Learning}
\label{alg:fermi6g_local}

\textbf{Inputs:} 
For each agent $i \in \mathcal{A}$: DRQN policy $\pi_i$ with parameters $\theta_i$, 
target network $\theta_i^{\text{target}}$, LSTM hidden state $h_i$, prioritized replay buffer $B_i$, optimizer, and ECDH key pair. \\
Global parameters: energy threshold $E_{\text{threshold}}$, reward weights ($w_L$, $w_E$, $w_F$, $w_{SE}$, $w_R$, $w_{EE}$, and $w_{MAC}$),  
number of episodes $N$, steps per episode $T$, and learning parameters $(batch\_size, \gamma, \epsilon, \delta, clip\_norm)$.

\begin{algorithmic}[1]
\STATE \textbf{Initialization:}
\FOR{each agent $i \in \mathcal{A}$}
    \STATE Initialize policy $\pi_i$ and set $\theta_i^{\text{target}} \gets \theta_i$
    \STATE Initialize hidden state $h_i \gets 0$, replay buffer $B_i$, and optimizer
    \STATE Generate ECDH key pair
    \STATE Set initial exploration rate $\epsilon \gets \epsilon_0$
\ENDFOR

\STATE \textbf{Training Loop:}
\FOR{episode $n = 1$ to $N$}
    \STATE Reset environment and agent states
    \FOR{each agent $i$}
        \STATE Set $h_i(0) \gets 0$
    \ENDFOR
    \FOR{timestep $t = 1$ to $T$}
        \FOR{each agent $i$}
            \STATE Observe state $s_i(t) \gets \texttt{env.get\_state}(i)$
            \STATE Encode features $\tilde{o}_i(t) \gets \texttt{observe}(s_i(t))$
            \STATE Update hidden state $h_i(t) \gets 
            \texttt{LSTM}(h_i(t-1), \tilde{o}_i(t))$
            \STATE Select action $a_i(t) \sim \pi_i(h_i(t), \epsilon)$
            \STATE Execute action: $(s_i(t{+}1), r_i(t), p_i(t), metrics) \gets \texttt{env.step}(i, a_i(t))$
            \STATE Store transition $(h_i(t), a_i(t), r_i(t), s_i(t{+}1), p_i(t))$ in $B_i$
            \STATE Sample minibatch from $B_i$ and compute TD loss
            \STATE Clip gradients and update $\theta_i$
        \ENDFOR
    \ENDFOR
    \STATE Update exploration rate: $\epsilon \gets \max(\epsilon_{\text{end}}, \epsilon \cdot \delta)$
\ENDFOR
\end{algorithmic}
\end{algorithm}

FERMI-6G Sub-Algorithm 1A captures the local reinforcement learning process at each agent, including state observation, action selection, experience replay, and prioritized updates. These updates are performed independently before global aggregation.

\begin{algorithm}[t]
\caption{FERMI-6G Sub-Algorithm 1B: Federated Aggregation \& Reward Adaptation}
\label{alg:fermi6g_global}

\textbf{Inputs:} Same as Sub-Algorithm 1A, plus aggregation interval $K$ and reward adaptation flag.

\begin{algorithmic}[1]
\STATE \textbf{Secure Aggregation Step (every $K$ episodes):}
\IF{$n \bmod K = 0$ and $n > 0$}
    \FOR{each agent $i \in \mathcal{A}$ with $E_i > E_{\text{threshold}}$}
        \STATE Generate ECDH key pair (if not existing)
        \STATE Compute AES masks $m_{ij}$ and aggregate:
               $M_i = \sum_{j>i} m_{ij} - \sum_{j<i} m_{ji}$
        \STATE Mask local model: $\theta_i' = \theta_i + M_i$
        \STATE Send $\theta_i'$ to aggregator
    \ENDFOR
    \STATE Aggregator computes global model:
           $\theta_g = \frac{1}{|\mathcal{A}|}\sum_i \theta_i'$
    \STATE Broadcast $\theta_g$ to all agents
    \STATE Each agent updates:
           $\theta_i \gets \theta_g$, $\theta_i^{\text{target}} \gets \theta_g$
\ENDIF

\STATE \textbf{Reward Adaptation:}
\IF{$reward\_adaptation\_enabled$}
    \STATE Adjust $(w_L, w_E, w_F)$
           based on system-level thresholds
    \STATE Normalize reward weights:
           $w_j \gets w_j / \sum_k w_k$
\ENDIF

\STATE \textbf{Logging:}
\STATE At episode end, record mean reward $\mathbb{E}[r_i]$, latency, energy,
       fairness index, and model divergence
\end{algorithmic}
\end{algorithm}

Once local training in FERMI-6G Sub-Algorithm 1A is complete, FERMI-6G Sub-Algorithm 1B executes, performing secure federated aggregation of model parameters and optional reward adaptation. This step ensures robustness against energy-constrained agents while optimizing fairness and efficiency. During aggregation, only agents meeting the energy threshold participate, and masking ensures privacy of local updates. Reward adaptation dynamically adjusts system-level weights to balance latency, energy, and fairness, maintaining overall performance across heterogeneous agents. Finally, relevant metrics are logged at the end of each episode to monitor convergence and guide future adaptation.

\subsection{System Integration and Optimizations}

Algorithm~\ref{alg:fermi6g_local} captures the core FERMI-6G mechanisms, linking each module to system-level components. Each agent uses an LSTM-based DRQN to encode temporal dependencies, producing actions across application, MAC, and CPU layers. These actions influence delay, interference, and energy metrics, which feed into both the reward computation and experience prioritization. Transitions with higher delay or interference are sampled more frequently to improve convergence.

Secure aggregation (Algorithm~\ref{alg:fermi6g_global}) employs ECDH/AES masking to protect model parameters, and agents below the energy threshold $E_{\text{threshold}}$ are excluded from updates. Reward adaptation (optional) adjusts weights $\alpha_j$ based on moving averages of system metrics to adapt in real time to latency, energy, and fairness variations. Experiments show that using fixed, pre-defined weights also achieves good performance, so this step can be optionally enabled depending on implementation preference.

\subsection{Key Innovations and Differentiators}

Unlike prior works that address individual layers in isolation~\cite{ref18,ref39}, FERMI-6G integrates privacy-preserving federated synchronization, energy-aware model filtering, and real-time reward adaptation into a single cross-layer learning framework. It jointly optimizes application, MAC, and CPU-layer actions while maintaining reliability under realistic 6G conditions, including dynamic spectrum, non-i.i.d. traffic, and resource-constrained edge devices.

\section{Experimental evaluation}
\label{sec:experimental_evaluation}
This section evaluates the performance of the proposed FERMI-6G (Fed-MARL) framework under various network conditions, using quantitative metrics. The evaluation compares FERMI-6G with multiple baselines to assess adaptability, scalability, robustness, and privacy resilience in ultra-dense 6G edge networks. The simulations were conducted on a MacBook Pro (13-inch, 2016), equipped with a 2.9 GHz Intel Core i5 processor, 8 GB of RAM, and Intel Iris Graphics 550. 

\subsection{Evaluation Framework and Metrics}
To evaluate FERMI-6G's performance, we use a set of key metrics that address various aspects of system effectiveness, including Reliability, Latency, Energy Consumption, Energy Efficiency, Spectral Efficiency, Fairness, MAC Success Rate, Task Completion Time, Offloading Delay, Failure Rate, Throughput, Scalability, and Privacy Protection. Task Completion Time also known as latency for successful tasks, measures the overall time required to process and complete a task, factoring in offloading and computation delays. Offloading Delay specifically tracks the time spent offloading tasks to the edge server, providing insights into network responsiveness. The Failure Rate monitors the percentage of tasks that fail due to issues like congestion or resource unavailability, with a lower failure rate indicating better robustness. Throughput quantifies the total data successfully transmitted over the network, evaluating how efficiently the system utilizes available resources. Scalability assesses the system's ability to handle a growing number of agents or higher network loads without significant performance degradation. Finally, Privacy Protection evaluates the system’s capacity to secure model updates through a secure aggregation protocol, ensuring privacy during the offloading process. These metrics collectively inform the Reward Function, which is designed to optimize the trade-off between latency, energy efficiency, fairness, spectral efficiency, and reliability. Together, they provide a comprehensive framework for assessing FERMI-6G's performance under dynamic, real-world conditions.

\subsection{Simulation Setup for Evaluating FERMI-6G in Dense Urban 6G Environments}
The simulation setup for evaluating FERMI-6G is designed to replicate the conditions of a dense urban 6G network, incorporating heterogeneous edge devices such as IoT devices, autonomous vehicles, and base stations, all operating in dynamic environments. The setup integrates multi-agent reinforcement learning (MARL) principles to manage tasks like offloading, energy management, and resource allocation across agents, ensuring that these components are responsive to the challenges of a 6G network.

\subsubsection{Network Model and Agent Configuration}
The network consists of a defined number of agents, each representing different devices like IoT sensors or vehicles. These agents are positioned randomly on a 100x100 grid, simulating the dynamic mobility of devices in a busy urban environment. Mobility patterns follow a random walk model, with agent velocities ranging from 0.1 to 1.0 m/s. This randomness mimics the unpredictable movements of devices in urban environments, adding realism to the simulation. This setup allows positions to be updated each timestep to reflect the agents' movements and adapt to the changing network topology. The central resource manager, or edge server, is located at the coordinates [50, 50] on the grid.

\subsubsection{Task Types and Offloading}
The tasks performed by the agents are categorized into three types, each with distinct characteristics to reflect real-world communication demands in 6G. URLLC tasks have a small size (1 MB) and a stringent deadline of 2 seconds, eMBB tasks are larger (3 MB) with a 5-second deadline, and mMTC tasks are smaller (0.5 MB) but have the longest deadline of 10 seconds. At the beginning of each simulation episode, agents are randomly assigned one of these task types, and their offloading strategies are influenced by factors such as local energy levels, CPU usage, and queue lengths.

\subsubsection{Learning Model}
Each agent is modeled as an independent learning entity, utilizing Deep Recurrent Q-Networks (DRQN) with Long Short-Term Memory (LSTM) layers. This setup allows agents to handle the partial observability of the environment, which is a consequence of the dynamic nature of the agents’ mobility, energy consumption, and other variables. The primary objectives of the agents are to manage task offloading to the edge server, optimize energy usage while maintaining battery levels above a 20\% threshold, and balance CPU usage to meet task deadlines. Each agent makes decisions based on local state observations, including task queue length, energy levels, CPU usage, channel availability, task type, and mobility speed, all of which impact their task offloading, energy management, and resource allocation strategies.

\subsubsection{Medium Access Control (MAC) Layer}
The network employs a medium access control (MAC) layer with a set number of channels, each with an initial bandwidth of 150 GHz and a channel capacity set to 2 per channel. The number of available channels is dynamic, and agents attempt to access these channels for task offloading and communication with the edge server. Channel congestion is modeled by tracking the number of agents using each channel, and retransmission probabilities are introduced to account for failed transmission attempts. The maximum queue length for each agent is set to 10, which can lead to retransmission if tasks fail.

\subsubsection{Simulation Parameters}
The simulation runs for 800 episodes with 20 steps per episode, during which agents interact with the environment to improve their task offloading and resource management strategies. Key experiment parameters include a 30\% probability of retransmitting failed offloading attempts, a noise standard deviation of 0.05 for modeling random noise during channel gain calculations, and a 2-second transmission delay per attempt. Energy consumption is modeled separately for task computation and transmission, and agents are penalized when their energy falls below the 20\% threshold.
The simulation setup also incorporates several additional parameters, summarized in the table below:

\begin{table}[htbp]
\centering
\caption{Simulation Parameters for the FERMI-6G Framework}
\resizebox{\columnwidth}{!}{
\begin{tabular}{|l|l|l|}
\hline
\textbf{Parameter} & \textbf{Description} & \textbf{Value/Range} \\
\hline
\texttt{NUM\_AGENTS} & Number of agents in the simulation & 5 \\
\texttt{EPISODES} & Total number of episodes for training & 800 \\
\texttt{STEPS} & Steps per episode & 20 \\
\texttt{HIDDEN} & Hidden layer size for agent networks & (2, 3, 3) \\
\texttt{BUFFER} & Size of the replay buffer & 10,000 \\
\texttt{BATCH} & Mini-batch size for training & 16 \\
\texttt{GAMMA} & Discount factor for future rewards & 0.99 \\
\texttt{LR} & Learning rate for the agent optimizer & 1e-3 \\
\texttt{EPS\_START} & Initial epsilon value for exploration-exploitation tradeoff & 1.0 \\
\texttt{EPS\_END} & Final epsilon value & 0.05 \\
\texttt{EPS\_DECAY} & Decay rate for epsilon value over episodes & 0.995 \\
\texttt{SYNC\_FREQ} & Frequency of synchronizing target networks & 10 \\
\texttt{SEQUENCE\_LENGTH} & Length of the sequence in DRQN model & 20 \\
\texttt{CLIP\_NORM} & Gradient clipping norm to prevent exploding gradients & 1.0 \\
\texttt{SMOOTHING\_WINDOW} & Window size for moving average smoothing of performance metrics & 10 \\
\hline
\end{tabular}
}
\label{table:sim_params}
\end{table}

As the simulation progresses, agents continuously update their strategies, refining them to maximize performance in terms of task completion, energy efficiency, and communication reliability.

\subsection{Framework Comparison and Results}

The experimental setup includes four frameworks for comparison: FERMI-6G, our proposed decentralized cross-layer approach; Fed-MARL, the same framework without cross-layer optimization; Centralized RL, a standard centralized RL with a single-layer policy; and Centralized RL (Cross-Layer), a centralized RL with full cross-layer optimization. All models were tested in environments that mimic realistic 6G network dynamics; however, due to the complexity of the simulated 6G network, simplified versions of the environment were used for the centralized RL frameworks to allow meaningful training and avoid extremely low rewards (which reached around -147) or zero reliability. Each framework was equipped with a prioritized replay buffer to isolate the effects of replay prioritization from the impact of cross-layer optimization. For clarity, the shorthand notation used throughout this paper is as follows: FERMI-6G refers to Fed-MARL with cross-layer optimization, Fed-MARL refers to Fed-MARL without cross-layer optimization, Centralized RL refers to standard centralized RL, and Centralized RL (Cross-Layer) refers to centralized RL with cross-layer optimization.

\begin{figure}[!t]
    \centering
    \vspace{-2ex}  
    \subfloat[]{%
        \includegraphics[width=0.5\linewidth,height=0.5\textheight,keepaspectratio]{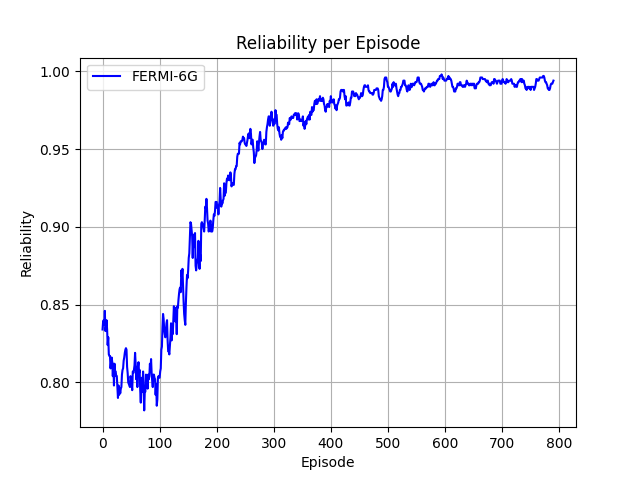}%
        \label{fig:subfig_a}}
    \hfill
    \subfloat[]{%
        \includegraphics[width=0.5\linewidth,height=0.5\textheight,keepaspectratio]{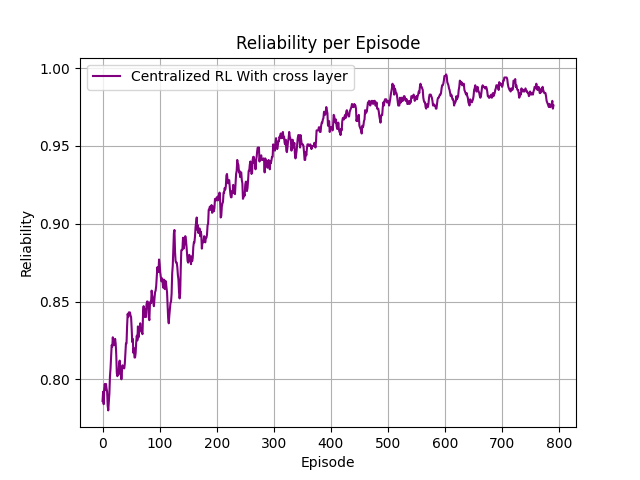}%
        \label{fig:subfig_b}} \\[-1ex]
    \vspace{-1ex}  
    \subfloat[]{%
        \includegraphics[width=0.5\linewidth,height=0.5\textheight,keepaspectratio]{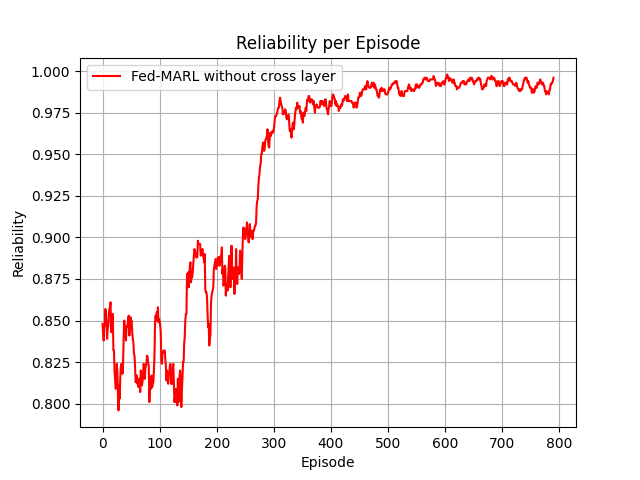}%
        \label{fig:subfig_c}}
    \hfill
    \subfloat[]{%
        \includegraphics[width=0.5\linewidth,height=0.5\textheight,keepaspectratio]{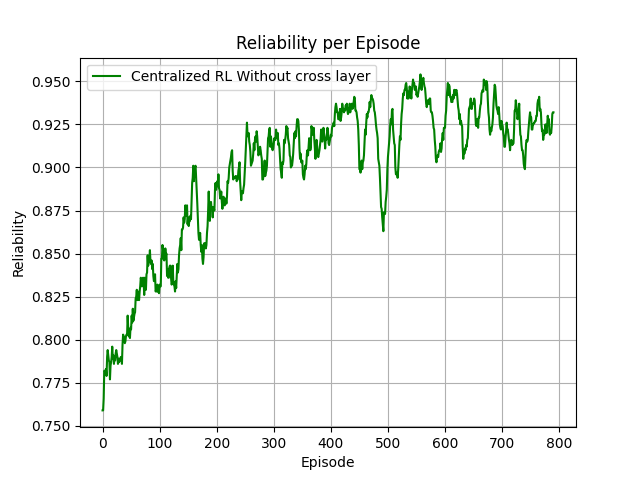}%
        \label{fig:subfig_d}}
    \caption{Reliability comparison across the four frameworks.}
    \label{fig:reliability_comparison}
\end{figure}

FERMI-6G incorporates cross-layer optimization, enabling agents to make decisions across multiple layers (application, MAC, CPU) based on local observations, thereby enhancing performance in complex environments. In contrast, Centralized RL uses Deep Recurrent Q-Networks (DRQN) with global state knowledge, but decisions are made using a simplified, single-layer policy, which limits its ability to handle complex, real-world scenarios. Centralized RL (with Cross-Layer) also uses DRQN but includes full cross-layer optimization, enabling global observability and decision-making across all layers. However, it is not realistic for real-world 6G systems due to privacy and communication constraints. To evaluate the impact of cross-layer optimization, we also tested Fed-MARL, which removes the cross-layer optimization while maintaining federated learning and secure aggregation. This version helps isolate the benefits of cross-layer coordination by comparing it to the FERMI-6G framework with cross-layer optimization. To illustrate the impact of the different frameworks on reliability, we present Fig. \ref{fig:reliability_comparison}, which shows the reliability performance across all experimental setups. Specifically, Fig. \hyperref[fig:subfig_a]{2(a)} shows the reliability of FERMI-6G, Fig. \hyperref[fig:subfig_b]{2(b)} shows Centralized RL with cross-layer optimization, Fig. \hyperref[fig:subfig_c]{2(c)} represents Fed-MARL, and Fig. \hyperref[fig:subfig_d]{2(d)} shows Centralized RL.

\paragraph{Baseline Refinement of Fed-MARL}
The initial baseline, Fed-MARL, performed poorly, with an average reward of -46 and reliability of only $\sim$30\%. This was due to stochastic application-level and MAC-layer actions, particularly the random MAC allocation that introduced high variance and instability. To enable meaningful comparison, we first removed the stochasticity in application and MAC decisions and introduced a deterministic round-robin MAC scheduler, where each channel sequentially grants access to agents. This ensured predictable and equitable channel allocation while preserving other network dynamics (task types, mobility, energy, and constraints). With this modification, Fed-MARL achieved improved reliability ($\sim$54\%) and more stable latency.

To further strengthen the baseline, we extended Fed-MARL so that each agent learns only application-level actions using a DRQN policy network with a single output head (\textit{app\_head}) for offloading decisions. MAC channel selection and CPU allocation are instead handled by fixed heuristics: the least-used channel is selected for MAC access, and CPU usage remains fixed at a default level. The environment continues to simulate full network dynamics (including channel contention, energy consumption, throughput, fairness, latency, reliability, and MAC success) and incorporates these into the reward signal. Agents interact independently, with no cross-layer coordination, while periodic secure federated aggregation (FedAvg) enables knowledge sharing. With these enhancements, Fed-MARL achieved a reliability of $\sim$94\%, establishing a robust baseline for fair comparison despite the absence of cross-layer learning.

\begin{table*}[htbp]
\centering
\caption{Framework Comparison (Mean ± Standard Deviation)}
\small  
\resizebox{\textwidth}{!}{
\begin{tabular}{|l|l|l|l|l|l|}
\hline
\textbf{Metric} & \textbf{Range} & \textbf{FERMI-6G} & \textbf{Centralized RL} & \textbf{Centralized RL (With Cross-Layer)} & \textbf{Fed-MARL} \\ \hline
\textbf{Reward} & env-dependent & 59.83 ± 6.70 & 51.58 ± 2.42 & 53.56 ± 2.97 & 51.51 ± 8.03 \\ \hline
\textbf{Reliability} & 0 → 100\% & 96.83\% ± 4.21\% & 89.78\% ± 5.32\% & 93.81\% ± 6.13\% & 94.00\% ± 7.31\% \\ \hline
\textbf{Energy Efficiency (bits/J)} & 0 → infinite & 68.72 ± 19.99 & 19.55 ± 0.71 & 19.84 ± 0.74 & 57.80 ± 16.37 \\ \hline
\textbf{Energy Consumption (J/task)} & 0 → (energymax=1) & 0.0230 ± 0.0075 & 0.0663 ± 0.0038 & 0.0683 ± 0.0040 & 0.0264 ± 0.0072 \\ \hline
\textbf{Spectral Efficiency (bps/Hz)} & 0 → env-dependent & 0.1981 ± 0.3750 & 0.5391 ± 0.0511 & 0.5642 ± 0.0575 & 0.2189 ± 0.4485 \\ \hline
\textbf{Latency (all attempts, s)} & 0 → infinite & 1.12 ± 0.69 & 1.36 ± 0.09 & 0.92 ± 0.40 & 1.11 ± 0.73 \\ \hline
\textbf{Task Completion Time (s)} & 0 → infinite & 0.95 ± 0.52 & 1.32 ± 0.63 & 0.72 ± 0.23 & 0.92 ± 0.53 \\ \hline
\textbf{Failure Rate (\%)} & 0 → infinite & 3.17\% ± 4.21\% & 10.22\% ± 5.32\% & 6.19\% ± 6.13\% & 5.99\% ± 7.31\% \\ \hline
\textbf{Privacy Protection} & - & High & Low & Low & High \\ \hline
\textbf{Offloading Delay (s)} & 0 → infinite & 3.95 ± 0.79 & 0.88 ± 0.76 & 0.70 ± 0.63 & 3.91 ± 0.78 \\ \hline
\textbf{Throughput (Gbps)} & 0 → env-dependent & 29.20 ± 2.75 & 11.35 ± 0.13 & 14.11 ± 0.14 & 28.26 ± 2.88 \\ \hline
\textbf{Scalability (Reliability with 50 Agents)} & 0 → 100\% & 90.01\% ± 0.60\% & 12.38\% ± 4.92\% & 20.69\% ± 9.10\% & 84.05\% ± 5.17\% \\ \hline
\textbf{Fairness} & 0 → 1 & 0.79 ± 0.07 & 0.76 ± 0.05 & 0.99 ± 0.0046 & 0.17 ± 0.07 \\ \hline
\textbf{MAC Success Rate (\%)} & 0 → 100\% & 96.83\% & 99.99\% & 99.99\% & 94.00\% \\ \hline
\end{tabular}
}
\label{table:frameworks}
\end{table*}

The results from our experimental setup are summarized in Table~\ref{table:frameworks}, which compares the four frameworks across several key performance metrics. In this section, we explain these metrics and their implications.

As shown in Fig.~\ref{fig:reliability_comparison}, the reliability performance of FERMI-6G and Centralized RL with Cross-Layer Optimization diverges notably during the early episodes. FERMI-6G begins with low reliability during episodes 0--100, where the reliability remains flat before gradually improving. This increases exponentially, stabilizing at 96.83\% $\pm$ 4.21\% by episode 800. In contrast, Centralized RL with Cross-Layer Optimization starts with a much higher initial reliability and improves more rapidly. It begins increasing almost immediately, stabilizing at 93.81\% $\pm$ 6.13\% by episode 400. This rapid early growth contrasts with FERMI-6G's slower initial progress, but ultimately, FERMI-6G reaches a higher final reliability. The initial delay in FERMI-6G's performance can be attributed to the time required for cross-layer optimization to adapt and deliver performance gains, whereas Centralized RL with Cross-Layer Optimization benefits from centralized visibility of the entire system from the outset, enabling quicker decisions that impact reliability. Fed-MARL, on the other hand, exhibits substantial reliability instability, fluctuating between 30\% and 70\% until episode 300. After that, it stabilizes at around 94\% $\pm$ 7.31\%, offering similar final reliability to Centralized RL with Cross-Layer Optimization but with more early-stage volatility. Centralized RL without cross-layer optimization demonstrates the lowest overall reliability, stabilizing at 89.78\% $\pm$ 5.32\%, with reliability oscillating between 87.0\% and 95.0\% during episodes 400--800.

\begin{figure}[!t]
    \centering
    \vspace{-2ex}  
    \subfloat[]{%
        \includegraphics[width=0.5\linewidth,height=0.5\textheight,keepaspectratio]{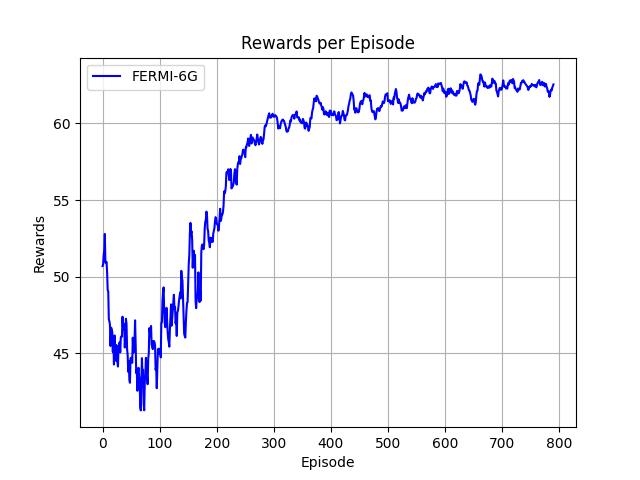}%
        \label{fig:subfig_a_rewards}}
    \hfill
    \subfloat[]{%
        \includegraphics[width=0.5\linewidth,height=0.5\textheight,keepaspectratio]{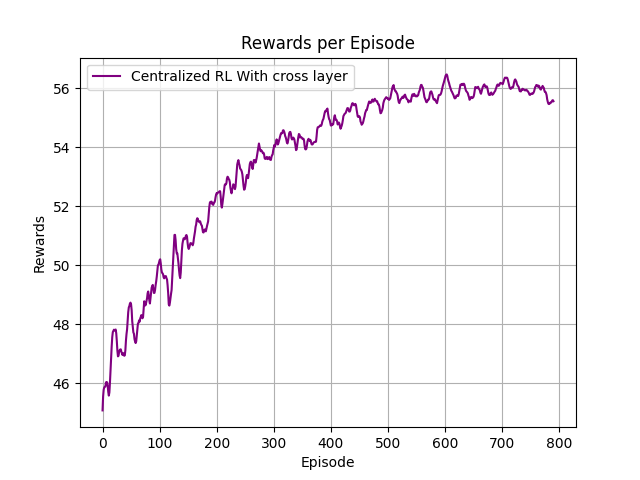}%
        \label{fig:subfig_b_rewards}} \\[-1ex]
    \vspace{-1ex}  
    \subfloat[]{%
        \includegraphics[width=0.5\linewidth,height=0.5\textheight,keepaspectratio]{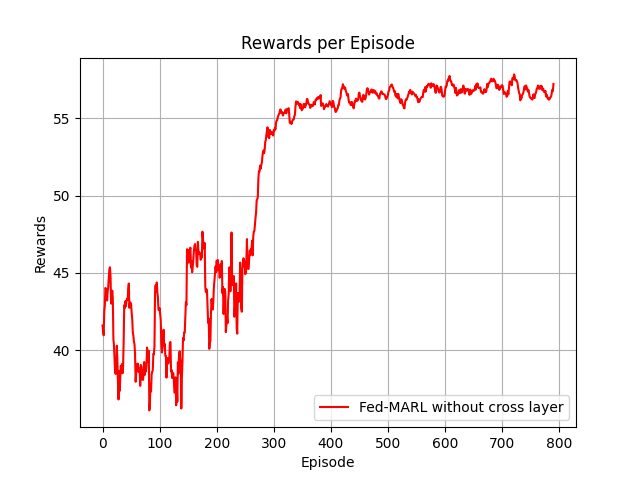}%
        \label{fig:subfig_c_rewards}}
    \hfill
    \subfloat[]{%
        \includegraphics[width=0.5\linewidth,height=0.5\textheight,keepaspectratio]{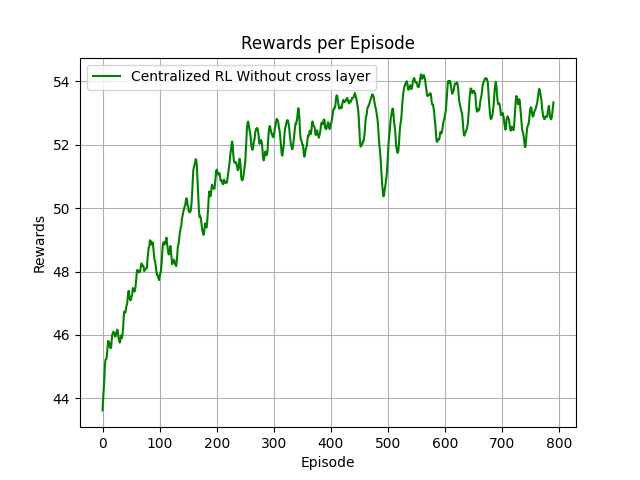}%
        \label{fig:subfig_d_rewards}}
    \caption{Rewards comparison across the four frameworks.}
    \label{fig:rewards_comparison}
\end{figure}

Fig.~\ref{fig:rewards_comparison} illustrates the reward dynamics across the frameworks. Centralized RL with Cross-Layer Optimization shows the most stable reward performance, reaching a steady value of 56. The rewards for FERMI-6G, however, are more erratic, with an initial dip to around 10 at episode 0, before recovering and stabilizing at 59.83 $\pm$ 6.70 by episode 800. Fed-MARL starts with significant reward instability, fluctuating between 10 and 48 rewards from episode 0 to 280, before stabilizing at 51.51 $\pm$ 8.03 after episode 280. This performance is less consistent than the other frameworks and can be attributed to the exploration phase, where the system learns optimal strategies. Centralized RL without cross-layer optimization experiences similar fluctuations, though it increases steadily to 54.31 $\pm$ 2.42 rewards in the early episodes, before fluctuating between 51 and 54 rewards in later episodes (after episode 400). This shows that although it benefits from an initial exponential increase, it struggles to stabilize in the long term compared to other frameworks.

\begin{figure}[!t]
    \centering
    \vspace{-2ex}  
    \subfloat[]{%
        \includegraphics[width=0.5\linewidth,height=0.5\textheight,keepaspectratio]{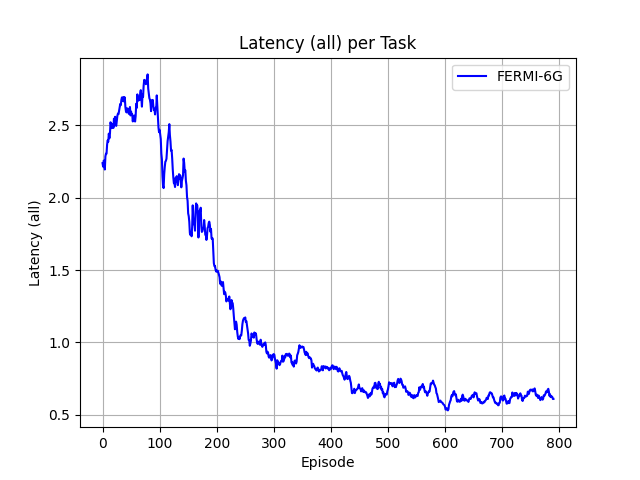}%
        \label{fig:subfig_a_latency}}
    \hfill
    \subfloat[]{%
        \includegraphics[width=0.5\linewidth,height=0.5\textheight,keepaspectratio]{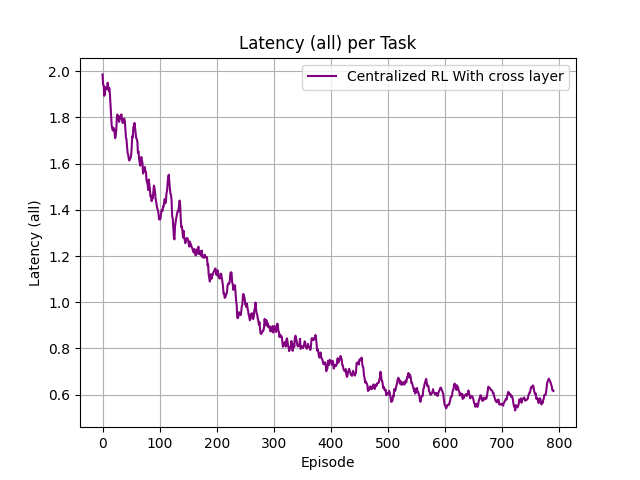}%
        \label{fig:subfig_b_latency}} \\[-1ex]
    \vspace{-1ex}  
    \subfloat[]{%
        \includegraphics[width=0.5\linewidth,height=0.5\textheight,keepaspectratio]{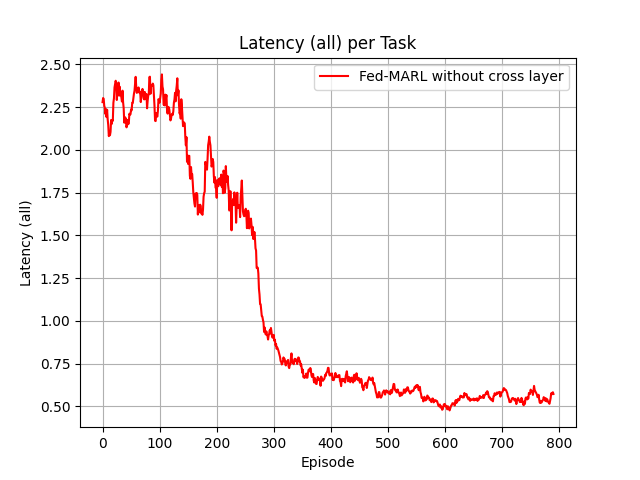}%
        \label{fig:subfig_c_latency}}
    \hfill
    \subfloat[]{%
        \includegraphics[width=0.5\linewidth,height=0.5\textheight,keepaspectratio]{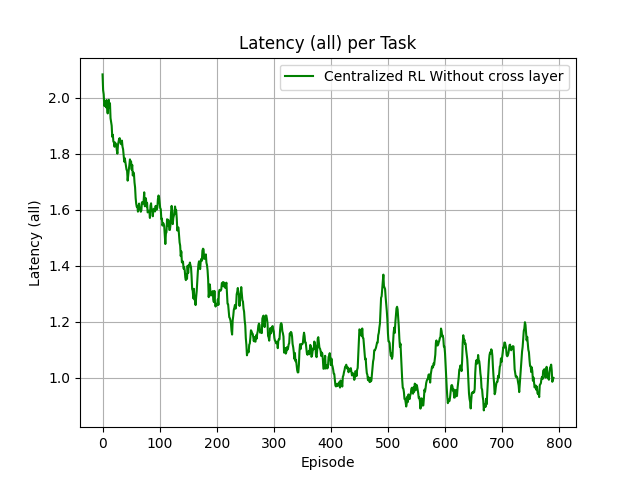}%
        \label{fig:subfig_d_latency}}
    \caption{Latency comparison across the four frameworks.}
    \label{fig:latency_comparison}
\end{figure}

Fig.~\ref{fig:latency_comparison} shows the latency performance, which includes both successful and failed task completions. FERMI-6G initially exhibits higher latency $\sim 2.8$ seconds at episode 100), but this improves exponentially and stabilizes at around 1.12 $\pm$ 0.69 seconds by episode 800. Centralized RL with Cross-Layer Optimization experiences the greatest reduction in latency, dropping from 2 seconds to below 0.6 seconds by episode 800. Its final latency is 0.92 $\pm$ 0.40 seconds, the lowest among all frameworks. Fed-MARL shows significant fluctuation, starting at 2.5--2.0 seconds in the first 150 episodes, but gradually decreasing to 1.72 seconds by episode 280. After that, it stabilizes but remains higher than FERMI-6G and Centralized RL with Cross-Layer Optimization, with a final value of 0.92 $\pm$ 0.53 seconds. Centralized RL without cross-layer optimization follows an erratic pattern. It initially drops quickly to around 0.4 seconds, but from episode 450 onward, the latency fluctuates between 1.4 and 0.4 seconds, ending at 1.36 $\pm$ 0.09 seconds by episode 800, the highest of all models.

\begin{figure}[!t]
    \centering
    \vspace{-2ex}  
    \subfloat[]{%
        \includegraphics[width=0.5\linewidth,height=0.5\textheight,keepaspectratio]{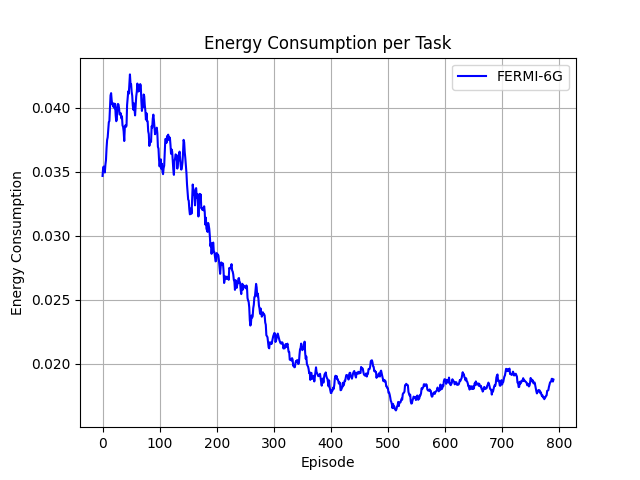}%
        \label{fig:subfig_a_energy}}
    \hfill
    \subfloat[]{%
        \includegraphics[width=0.5\linewidth,height=0.5\textheight,keepaspectratio]{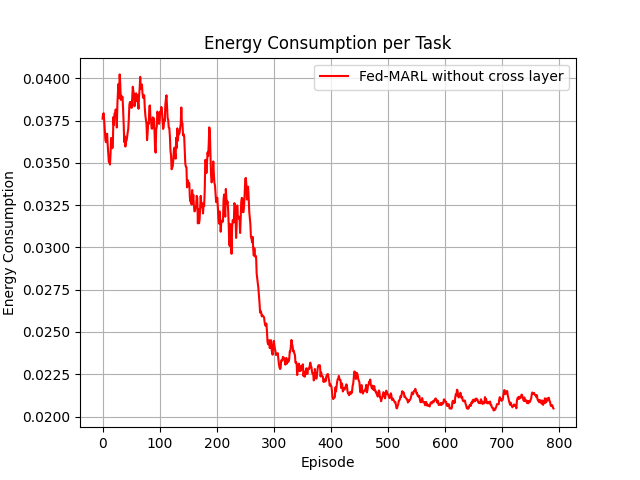}%
        \label{fig:subfig_b_energy}} \\[1ex]
    \vspace{-2ex}  
    \subfloat[]{%
        \includegraphics[width=0.5\linewidth,height=0.5\textheight,keepaspectratio]{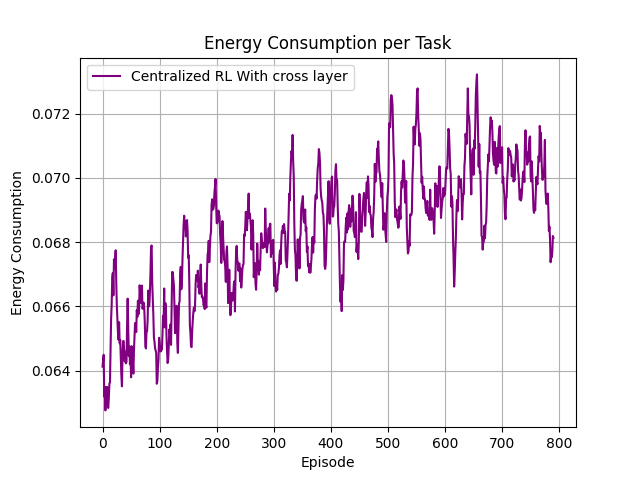}%
        \label{fig:subfig_c_energy}}
    \hfill
    \subfloat[]{%
        \includegraphics[width=0.5\linewidth,height=0.5\textheight,keepaspectratio]{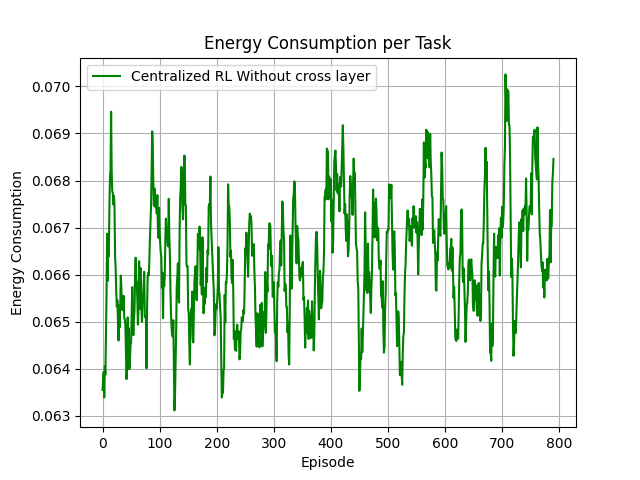}%
        \label{fig:subfig_d_energy}}
    \caption{Energy consumption comparison across the four frameworks.}
    \label{fig:energy_comparison}
\end{figure}

Fig.~\ref{fig:energy_comparison} illustrates the energy efficiency and consumption across the different frameworks. FERMI-6G leads in energy efficiency with 68.72 ± 19.99 bits/J, outperforming Centralized RL (19.55 ± 0.71 bits/J) and Centralized RL with Cross-Layer Optimization (19.84 ± 0.74 bits/J). In terms of energy consumption per task, FERMI-6G uses the least energy, at 0.0230 ± 0.0075 J/task, followed closely by Fed-MARL (0.0264 ± 0.0072 J/task). In contrast, both centralized approaches consume significantly more energy (0.0663 ± 0.0038 J/task for Centralized RL and 0.0683 ± 0.0040 J/task for Centralized RL with Cross-Layer Optimization), reflecting their higher computational demands.
Regarding energy consumption per agent, FERMI-6G starts at around 0.035 energy at episode 0, with fluctuations between 0.037 and 0.039. By episode 150, consumption begins to decrease steadily, reaching 0.011 by episode 800. Fed-MARL exhibits a similar trend but with slightly more fluctuation early on. It stabilizes at a lower value around episode 500, showing a marked reduction in energy consumption compared to earlier stages.

In contrast, both Centralized RL and Centralized RL with Cross-Layer Optimization exhibit unstable energy consumption. While Centralized RL with Cross-Layer maintains a somewhat consistent fluctuation, its overall energy consumption increases over time, with no clear downward trend. Centralized RL Optimization, despite offering full observability, still struggles with high and erratic energy consumption, lacking the optimization seen in decentralized frameworks like FERMI-6G and Fed-MARL. The absence of effective decentralized decision-making and resource optimization leads to inefficiencies, making these centralized approaches less energy-efficient.
As illustrated in Fig. \hyperref[fig:subfig_a_energy]{5(a)}, FERMI-6G demonstrates a significant reduction in energy consumption over time, from around 0.035 J/task at episode 0 to 0.011 J/task by episode 800, representing a 70\% decrease. This decline reflects the framework’s ability to optimize resource allocation across the application, MAC, and CPU layers, minimizing redundant computations. Fed-MARL follows a similar, though slightly less efficient, path, stabilizing at a higher energy consumption rate by episode 800.

Beyond reliability, rewards, and latency, the table also includes other metrics such as task completion time, failure rate, privacy protection, throughput, and scalability. For example, FERMI-6G exhibits the fastest task completion time (0.95 ± 0.52 seconds) and the lowest failure rate (3.17\% ± 4.21\%), compared to the other frameworks. These results are consistent with its superior performance in cross-layer optimization, which enables better task scheduling and resource allocation.
In terms of scalability, FERMI-6G outperforms all other frameworks, maintaining high reliability (90.01\% ± 0.60\%) even when scaled to 50 agents. In contrast, Centralized RL without cross-layer optimization faces severe scalability issues, with reliability dropping to just 12.38\% ± 4.92\% as 50 agents are added.
Finally, privacy protection is a critical feature of federated learning-based frameworks. FERMI-6G and Fed-MARL maintain high levels of privacy protection, whereas Centralized RL and Centralized RL with Cross-Layer Optimization exhibit lower levels due to their reliance on centralized control, which compromises user data privacy.

\section{Discussion}
\label{sec:discussion}
The comparison reveals key trade-offs between the different frameworks. FERMI-6G consistently outperforms the strengthened Fed-MARL baseline in terms of cumulative reward. Notably, as training episodes increase, FERMI-6G even surpasses the centralized cross-layer framework in application-level performance, highlighting the effectiveness of decentralized cross-layer optimization. However, this performance gain comes at the expense of fairness, while the centralized cross-layer framework maintains a fairness index of $\sim$0.80, FERMI-6G’s fairness drops significantly to 0.3734 ($\approx$37.34\%). This suggests that for FERMI-6G to maximize performance, agents must adopt more “selfish” policies, prioritizing their own gains over system-wide equity. This trade-off underscores a fundamental tension in decentralized multi-agent systems, while cross-layer coordination enhances efficiency and throughput, it may simultaneously erode fairness across agents. Understanding and mitigating this trade-off is essential for designing practical and equitable 6G systems.
The FERMI-6G and centralized RL cross-layer environments differ significantly in terms of realism and modeling detail. FERMI-6G provides a rich per-agent state representation, capturing queue lengths, energy levels, CPU usage, channel conditions with path loss, fading, and blockage, as well as contextual features including task type, size, deadline, mobility, device type, and battery health. MAC access and retransmissions are modeled realistically, accounting for congestion, collisions, and hybrid fairness metrics, with nuanced reward shaping that balances latency, energy, fairness, spectral efficiency, reliability, and MAC success using task-specific weights and dynamic penalties. By contrast, the centralized RL cross-layer environment simplifies the state and MAC modeling, employing limited channel and task context, a probabilistic collision model, and a basic reward function without task-specific differentiation or dynamic penalties. FERMI-6G also incorporates additional mechanisms such as edge caching and smoothed spectral efficiency, absent in the cross-layer model. Consequently, direct quantitative comparisons between the two frameworks may be misleading, as performance differences partly reflect the underlying environment assumptions rather than the learning strategies alone.
The spectral efficiency values reported here are normalized to a minimum threshold of 0.01 bps/Hz to ensure consistent performance in challenging environments. While 6G networks are expected to achieve spectral efficiency values in the range of 10-20 bps/Hz in ideal conditions, the lower values in our results reflect the fact that the models used here focus on general performance across varying network conditions. When using the full simulated 6G network, centralized RL frameworks experienced extremely low rewards (around -147) and zero reliability, indicating that standard centralized approaches, even with cross-layer optimization, struggle to handle the complexity of realistic 6G network dynamics. To enable meaningful training, the environments for centralized RL and centralized RL (with cross-layer) were simplified, reducing stochasticity in MAC access, queueing, and task execution. Despite these simplifications, reliability and fairness remained limited, and throughput and spectral efficiency metrics were less stable compared to FERMI-6G. These results emphasize that while centralized cross-layer learning can provide insights under controlled conditions, decentralized cross-layer approaches like FERMI-6G are more robust and effective in complex, realistic 6G scenarios.
These findings also highlight the relative advantages of modular heuristics versus learned cross-layer policies. Comparatively, in scenarios with moderate network dynamics, Fed-MARL can perform reasonably well because the heuristics handle MAC and CPU allocation effectively, while the learned application-level policy still makes competent offloading decisions. However, in highly dynamic or congested networks, FERMI-6G outperforms Fed-MARL, as its learned cross-layer policies can exploit adaptive CPU scaling and channel selection to optimize end-to-end performance. Regarding training cost, Fed-MARL is simpler, faster to train, and more robust to hyperparameter choices, whereas FERMI-6G requires careful tuning to avoid instability.
Finally, we note that our training procedure assumes full synchronization every K episodes. While this simplifies analysis, it may not fully reflect practical mobile 6G conditions where agents can be delayed or fail. Importantly, the secure aggregation protocol used in our framework naturally supports partial participation, enabling asynchronous aggregation without compromising security. This capability makes FERMI-6G more amenable to realistic deployment scenarios, where strict synchronization cannot always be guaranteed.

\section{Conclusion}
\label{sec:conclusion}

In this work, we introduced FERMI-6G, a federated multi-agent reinforcement learning framework for decentralized, privacy-preserving, and energy-aware resource management in ultra-dense 6G edge networks. By combining temporal deep reinforcement learning, a cross-layer composite action space, secure federated synchronization, and adaptive multi-objective optimization, FERMI-6G enables agents to jointly optimize task offloading, spectrum access, and CPU scaling under partial observability and mobility constraints. Experimental results showed that FERMI-6G surpasses existing baselines in reliability, latency, energy efficiency, and scalability while preserving privacy. However, challenges remain in enhancing fairness, reducing latency to meet URLLC requirements, and scaling training efficiency. As future work, we plan to improve spectral efficiency by expanding channel bandwidth, increasing SINR through advanced power control, and integrating technologies such as massive MIMO and refined interference management. We also aim to incorporate ultra-low latency techniques, including edge-native computing, efficient channel access, and integrated air–ground networks, to support demanding applications like immersive AR/VR, industrial IoT, and holographic communication. Finally, we will investigate fairness-aware coordination mechanisms, scalable federated learning strategies, and real-world testbed validation to bring FERMI-6G closer to the ambitious throughput, latency, and reliability targets of next-generation 6G systems.


\end{document}